\documentclass[10pt,twocolumn,letterpaper]{article}

\usepackage[pagenumbers]{iccv} 

\definecolor{iccvblue}{rgb}{0.21,0.49,0.74}
\usepackage[pagebackref,breaklinks,colorlinks,allcolors=iccvblue]{hyperref}
\usepackage{multirow} 
\usepackage{algorithm,algpseudocode}
\usepackage{colortbl} 

\title{Memory-efficient Low-latency Remote Photoplethysmography through Temporal-Spatial State Space Duality}

\author{Kegang Wang$^{*}$\\Tsinghua University\\
\and
Jiankai Tang$^{*}$\\Tsinghua University\\
\and
Yuxuan Fan\\Tsinghua University\\
\and
Jiatong Ji\\Tsinghua University\\
\and
Yuanchun Shi\\Tsinghua University\\
\and
Yuntao Wang$^{\dagger}$\\
Tsinghua University\\
}
\begin{document}

\maketitle

\begin{abstract}
    Remote photoplethysmography (rPPG), enabling non-contact physiological monitoring through facial light reflection analysis, faces critical computational bottlenecks as deep learning introduces performance gains at the cost of prohibitive resource demands. This paper proposes ME-rPPG, a memory-efficient algorithm built on temporal-spatial state space duality, which resolves the trilemma of model scalability, cross-dataset generalization, and real-time constraints. Leveraging a transferable state space, ME-rPPG efficiently captures subtle periodic variations across facial frames while maintaining minimal computational overhead, enabling training on extended video sequences and supporting low-latency inference. Achieving cross-dataset MAEs of 5.38 (MMPD), 0.70 (VitalVideo), and 0.25 (PURE), ME-rPPG outperforms all baselines with improvements ranging from 21.3\% to 60.2\%. Our solution enables real-time inference with only 3.6 MB memory usage and 9.46 ms latency -- surpassing existing methods by 19.5\%-49.7\% accuracy and 43.2\% user satisfaction gains in real-world deployments. The code and demos are released for reproducibility on \href{https://health-hci-group.github.io/ME-rPPG-demo/}{https://health-hci-group.github.io/ME-rPPG-demo/}.

\end{abstract}  

\section{Introduction}
\label{sec:intro}

Remote photoplethysmography (rPPG) is a non-contact technique that extracts physiological signals such as PPG from human skin by analyzing the light reflected from the facial skin surface~\cite{mcduff2015survey}. These subtle cyclic variations in the skin color may not be visible to the naked eye, but they can be captured by RGB cameras and analyzed to estimate the heart rate~\cite{wu2012eulerian}. The promising development of rPPG technology has enabled a wide range of applications, such as health monitoring~\cite{lu2021dual,liu2024summit}, emotion recognition~\cite{wang2023physbench,song_pulsegan_2020}, and driver fatigue detection~\cite{du2022multimodal}.

\begin{figure}[t]
    \centering
    \includegraphics[width=1.0\linewidth]{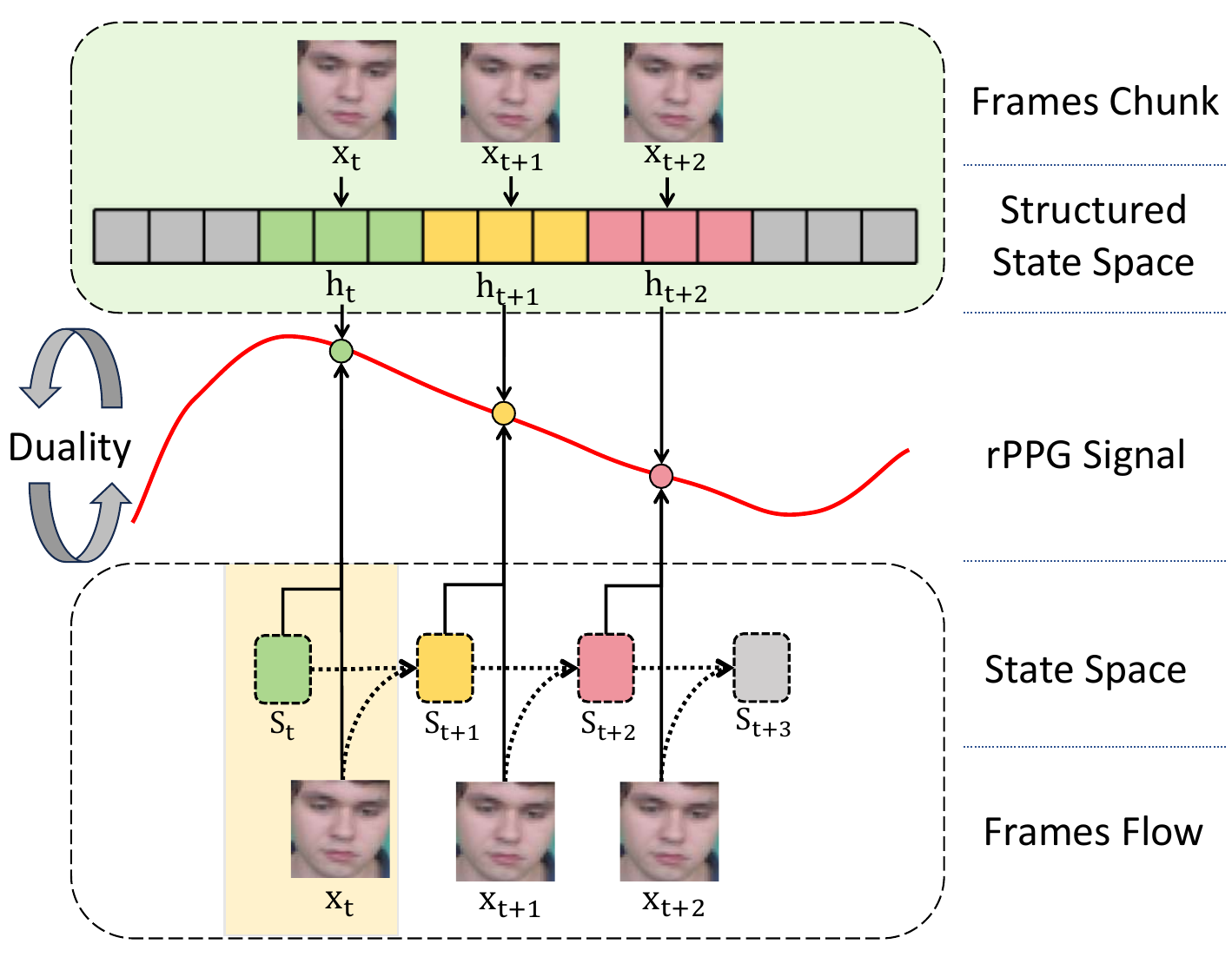}
    \caption{\textbf{The principle of state space duality.} The state space is transferable and aligns with each input frame. SSD models trained on frame chunks can generalize to single-frame inference. The model takes a single frame as input and outputs the corresponding state along with the rPPG prediction.}
    \label{fig:principle}
\end{figure}

The evolution of rPPG methods has progressed from traditional methods based on time-domain and frequency-domain analysis to deep learning-based methods~\cite{mcduff2023camera}. Deep learning methods have shown superior performance in rPPG estimation, especially in noisy environments~\cite{tang2023mmpd,nowara_benefit_2020}. However, the computational resources required by deep learning methods have also increased, making large-scale rPPG training and real-time inference challenging. Most algorithms are validated on small-scale datasets such as PURE~\cite{pure} and UBFC~\cite{ubfcrppg}, which may have data distribution biases and cannot represent the challenges in real-world scenarios. Efficient training and large-scale validation remain critical challenges in rPPG research.

Due to memory limitations in video loading and algorithm processing~\cite{he2019device,lin2019tsm}, rPPG algorithms typically require slicing videos into shorter segments (e.g., 180 frames, approximately 6 seconds) for batch training~\cite{liu2022rppgtoolbox,kuang2023shuffle}. This approach may overlook long-term temporal correlations, leading to the loss of important information during training, and making algorithms more susceptible to noise.

Similarly, rPPG models demand significant computational resources during inference~\cite{mttscan}. Some studies attempt to address efficiency issues by reducing the resolution of frames~\cite{rtrppg} or simplifying the model complexity~\cite{efficientphys}, but no method has successfully used a single frame to predict the rPPG signal so far. Most algorithms perform inference by slicing long videos into windows after recording, which limits the real-time performance of rPPG and its application on edge devices. In real-world scenarios, obtaining the rPPG output may take 10 seconds or longer (including recording, preprocessing, model inference, and postprocessing), which is not user-friendly or practical for medical applications~\cite{poh2011medical}. Therefore, achieving real-time rPPG waveform output during recording is particularly important, and this is the research goal of this paper.

This paper proposes a \textbf{M}emory-\textbf{E}fficient rPPG (\textbf{ME-rPPG}) built on \textbf{T}emporal-\textbf{S}patial \textbf{S}tate \textbf{S}pace \textbf{D}uality (\textbf{TSD}). ME-rPPG captures subtle temporal color changes and spatial skin representations corresponding to physiological signals, enabling accurate prediction of model states and blood volume pulse (BVP) values across arbitrary-length frames.
Leveraging state space duality (SSD), ME-rPPG can be trained on chunked frames and perform inference on a single frame, significantly reducing computational load and achieving real-time inference. ME-rPPG demonstrates superior performance in intra-dataset, cross-dataset, and real-world evaluation compared to state-of-the-art (SOTA) methods. The three core contributions of this paper are as follows: 
\begin{enumerate}
    \item We propose a memory-efficient rPPG algorithm based on temporal-spatial state space duality, which captures frame change trends and facial vascular information to extract temporal-spatial features, improving accuracy by 50.2\% on the challenging MMPD dataset.
    \item We evaluated ME-rPPG on large-scale datasets and demonstrated that it outperforms state-of-the-art methods, with improvements of up to 60.3\% in accuracy and 98.4\% in memory efficiency.
    \item To the best of our knowledge, ME-rPPG is the first rPPG model capable of real-time inference using only a single frame as input, requiring only 3.6 MB of memory. Real-world studies show that ME-rPPG has a better user experience, improving user satisfaction by 43.2\%. 
\end{enumerate}

\section{Related Work}
\label{sec:related}

\begin{figure}[t]
    \centering
    \includegraphics[width=1.0\linewidth]{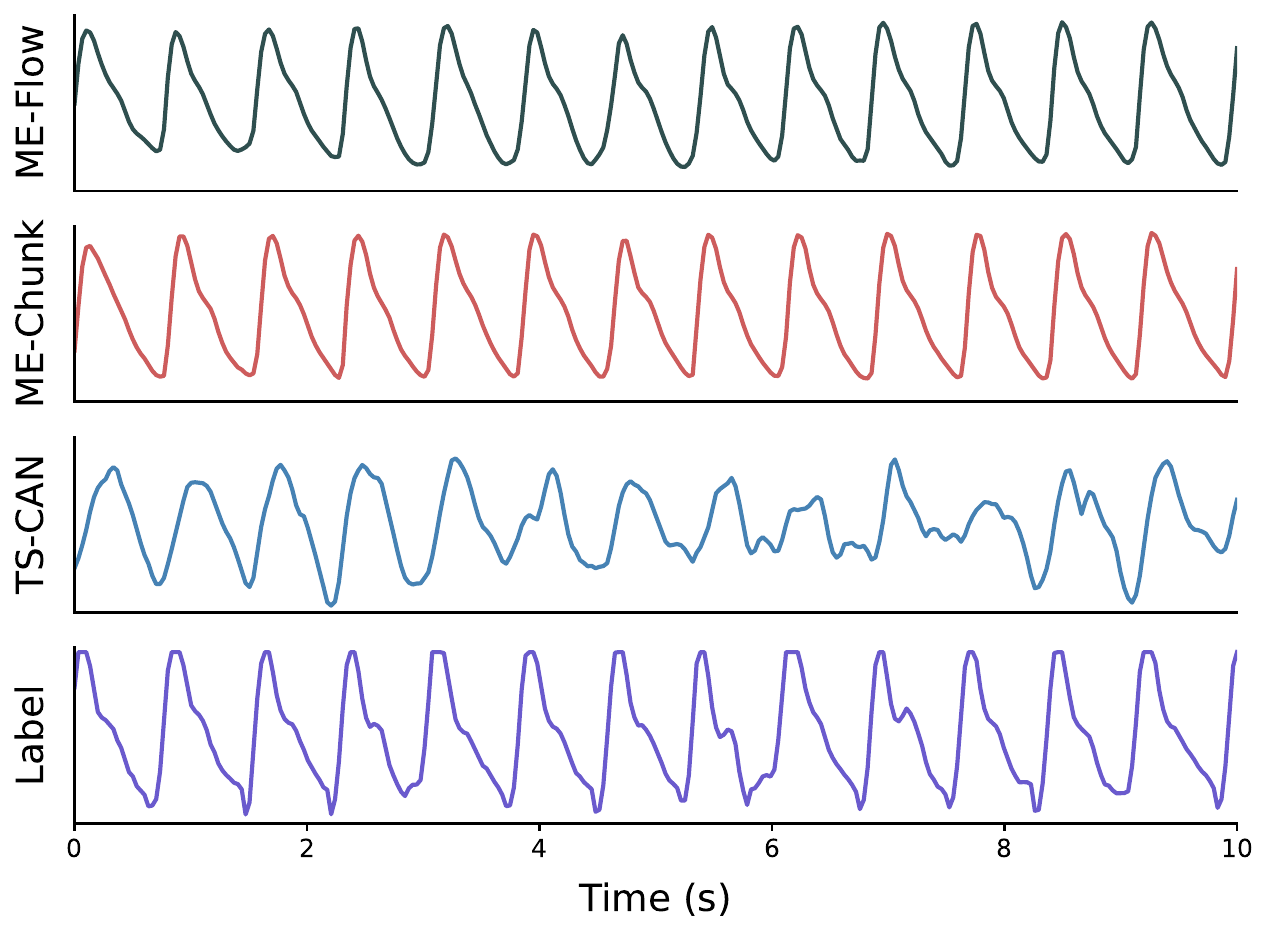}
\caption{\textbf{Comparison of Predictions and Ground Truth.} The predictions are obtained using ME-Flow, ME-Chunk, and TSCAN from a video in the VitalVideo dataset.}
    \label{fig:waveform}
\end{figure}

\label{sec:related:ppg}
\textbf{Efficient rPPG Methods.} Recent advances in rPPG signal extraction have shifted from traditional signal processing techniques, such as POS~\cite{pos}, to deep learning methods for rPPG signal extraction~\cite{Gideon2021TheWT,Li2023ContactlessPE}. These deep learning methods have significantly improved accuracy and robustness. Early deep learning models in rPPG relied on CNN-based architectures, leveraging their success in image processing tasks. These methods extract features from facial video frames using convolutional and pooling layers, and then predict the rPPG signal using fully connected layers. Various types of CNNs, including 1D-CNN, 2D-CNN, and 3D-CNN, have been explored. Seq-rPPG~\cite{wang2023physbench} focuses on temporal features of facial videos using 1D-CNN, achieving stable performance with only an 8x8 facial region. DeepPhys~\cite{chen2018deepphys} and TS-CAN~\cite{liu2020multi} use 2D-CNNs to extract spatial features and introduce temporal shift modules for better heart rate estimation. PhysNet~\cite{physnet} and iBVPNet~\cite{joshi2024ibvp} employ 3D-CNNs to extract spatio-temporal features, enhancing both accuracy and inference speed through attention mechanisms and lightweight networks. Despite their advancements, CNN-based models struggle to capture long-range dependencies, limiting their effectiveness in complex real-world scenarios. Transformer-based methods address this limitation by leveraging self-attention mechanisms to extract features from facial video frames. PhysFormer~\cite{yu2022physformer} is the first Transformer-based rPPG method, focusing on the temporal features of facial videos and achieving higher performance through long-range dependency self-attention mechanisms. RhythmFormer~\cite{Zou2024RhythmFormerER} proposes a hierarchical periodicity Transformer, achieving higher performance. Spiking-PhysFormer~\cite{Liu2024SpikingPhysFormerCR} proposes an rPPG method based on spiking neural networks, which achieves higher performance. Although Transformer methods have made significant progress in the rPPG field, their computational complexity makes them difficult to apply widely in practice. Therefore, more efficient methods are needed to extract rPPG signals.

\textbf{State Space Model.} State Space Models (SSMs) represent a class of mathematical models designed to overcome the quadratic memory constraints of Transformer architectures. Unlike Transformers, SSMs offer linear-time complexity for processing long sequences, making them computationally efficient for sequential data. Mamba \cite{gu2023mamba}, a selective state space model, effectively addresses Transformer's inefficiency on long sequences while maintaining competitive performance. The advanced version, Mamba-2, further bridges SSMs and Transformers through a Structured State Space Duality (SSSD) framework, enhancing computational efficiency \cite{dao2024transformers}. In the video understanding domain, VideoMamba efficiently handles long-term dependencies and high-resolution videos, significantly outperforming traditional feature models \cite{li2024videomamba}. Beyond video applications, Mamba models have demonstrated success across diverse visual applications including image processing, point cloud analysis, and multimodal data integration \cite{xu2024visual}. These advantages make SSMs a promising approach for real-time video processing, motivating their application to rPPG signal extraction.

\textbf{Mamba-Based rPPG Models.} Mamba-based methods have recently gained attention for rPPG extraction due to their efficiency in handling long-term dependencies. Recent studies have primarily utilized Mamba as a feature extractor, leveraging its state-space modeling capabilities for improved temporal representation. Unlike Transformer-based models, which rely on self-attention mechanisms, Mamba employs lightweight sequential modeling, making it well-suited for large-scale training and low-latency deployment in rPPG applications. For instance, PhysMamba\cite{luo2024physmamba} integrates a slow-fast temporal difference mechanism to enhance physiological signal extraction, while CardiacMamba\cite{wu2025cardiacmamba} fuses RGB and RF signals through a multimodal framework, improving measurement accuracy. Similarly, RhythmMamba~\cite{zou2024rhythmmamba} focuses on handling arbitrary-length videos, offering a flexible approach to physiological monitoring. Despite these advancements, existing Mamba-based methods primarily focus on feature extraction rather than fully utilizing state-space properties for efficient batch training and real-time inference. To date, Mamba’s full potential for optimizing rPPG models at scale remains underexplored.

\begin{table}[!t]
\centering
\caption{Computational Efficiency Comparison}
\label{tab:efficiency}
\begin{tabular}{lcccc}
\toprule
Method & Params & Mem. & Latency & FLOPs \\
\midrule
EFFPhys & 2,189 & 609 & 371 & 39.0 \\
TSCAN & 533 & 228 & 55.2 & 8.07 \\
PhysNet & 770 & 230 & 62.1 & 6.16 \\
PhysFormer & 7,395 & 576 & 2330 & 95.7 \\
RhyMamba & 4,936 & 347 & 889 & 27.4 \\
ME-Chunk & 580 & 260 & 526 & 20.6 \\
\rowcolor{gray!30}
ME-Flow & 580 &3.6 & 9.46 & 0.15 \\
\bottomrule
\end{tabular}
\scriptsize \noindent \\Params (K) = Number of parameters. Mem (MB) = Memory usage. Latency (ms) = Inference time on CPUs. FLOPs (G) = Floating point operations. ME-Chunk = ME-rPPG chunk inference. ME-Flow = ME-rPPG frame inference.
\end{table}
\section{Method}
\label{sec:method}

\subsection{General Framework}
\begin{figure*}[t]
    \centering
    \includegraphics[width=1.0\textwidth]{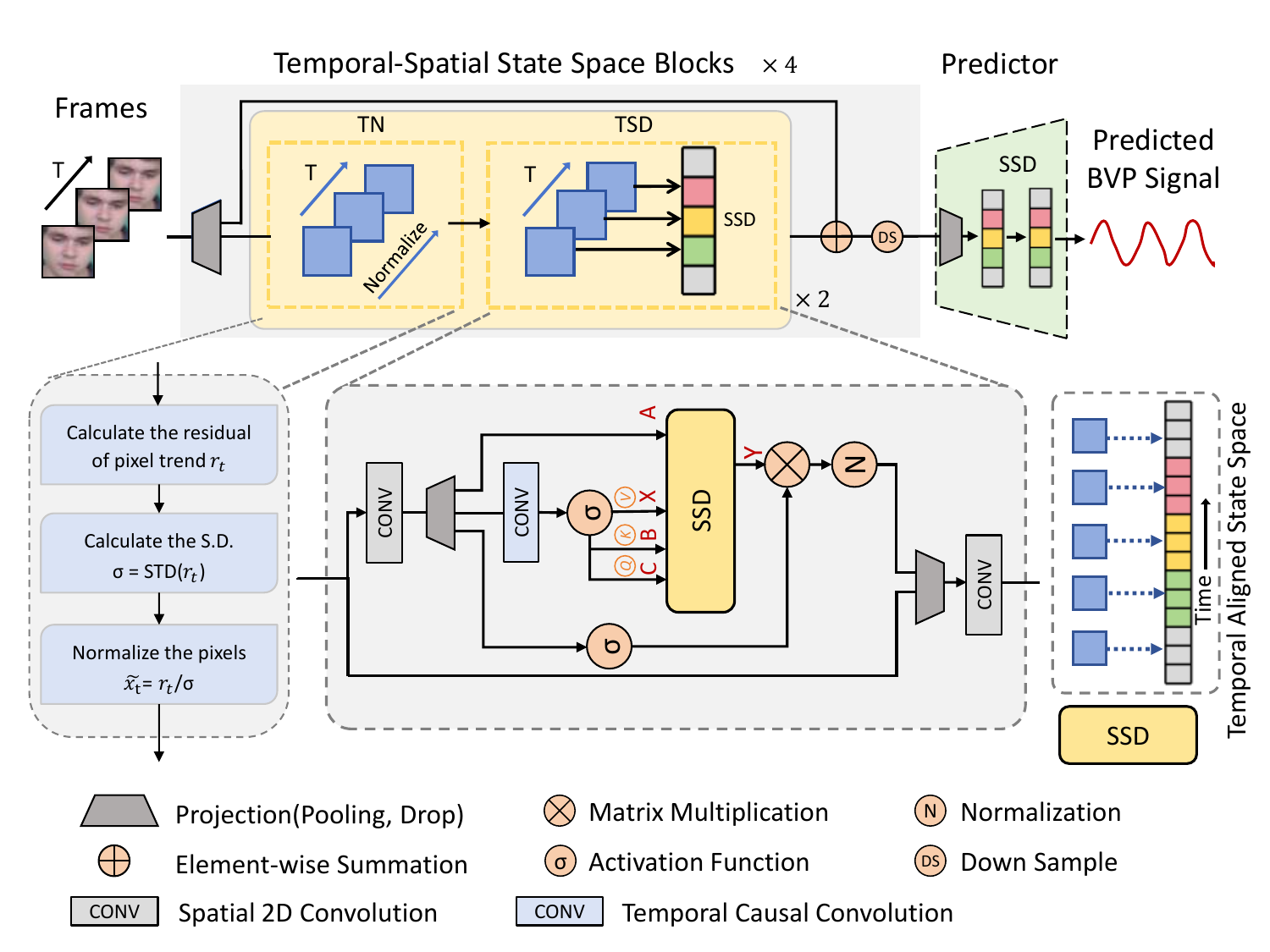}
    \caption{\textbf{General framework of ME-rPPG.} Our method takes resized facial frames as input and predicts a BVP value and a state. The TN module captures temporal variance while the TSD module extracts temporal-spatial features.}
    \label{fig:model}
\end{figure*}

Our framework processes video sequence tensors $\mathbf{X} \in \mathbb{R}^{B \times T \times H \times W \times C}$ as input and extracts blood volume pulse (BVP) signals $\mathbf{y} \in \mathbb{R}^{B \times T}$ through \textbf{T}emporal-\textbf{S}patial \textbf{S}tate \textbf{S}pace \textbf{B}locks (TSB). As illustrated in Figure~\ref{fig:model}, the TSB computational workflow comprises two critical stages:
\begin{enumerate}
    \item Temporal Normalization: Independently performs detrending and standardization across the temporal dimension for each spatial coordinate.
    \item Temporal-Spatial Modeling: Achieves long-range temporal dependency modeling through TSD.
\end{enumerate}
Unlike traditional spatio-temporal joint modeling, TSB decouples spatial and temporal dimensions, utilizing  2D convolutions for spatial modeling while employing TN and TSD for temporal modeling. This separation enables memory-efficient, step-by-step inference, ensuring temporal alignment between input and output. Consequently, the model can perform inference incrementally over time, significantly improving memory efficiency.
For a detailed comparison of memory efficiency with conventional architectures, refer to Table \ref{tab:memory complexity}.

\begin{table}[htbp]
  \centering
  \caption{Comparison of Memory Efficiency Between TSD and Traditional Structures.}
  \label{tab:memory complexity}
  \begin{tabular}{ccc}
    \toprule
    Structure       & Training Mem. & Inference Mem.          \\
    \midrule
    3D CNN      &  $O(Tk^3D)$   & $O(k^3D)$ \\
    Transformer      & $O(T^2D)$         & $O(TD)$     \\
    TSD (Ours)      & $O(TD)$         & $O(D+L)$     \\
    \bottomrule
  \end{tabular}
\scriptsize \noindent \\ $L$ = SSD state dim. $k$ = 3D kernel size. $D$ = Feature dim. $T$ = Time duration.
\end{table}

\subsection{Temporal Normalization Module}

The TN module\cite{wang2024plugandplaytemporalnormalizationmodule} ensures the stability of temporal characteristics in long sequences. It begins by calculating the linear trend term using the least squares method, as formulated in Equation \eqref{eq:detrend}. During training, the trend estimation can be computed in parallel.

\begin{equation}
(\beta_0, \beta_1) = \arg\min{\beta} \sum_{t=1}^T \left( x_t - (\beta_0 t + \beta_1) \right)^2 \label{eq:detrend}
\end{equation}

After obtaining the trend coefficients $(\beta_0, \beta_1)$, the residual of each pixel relative to the trend is computed and subsequently standardized as the final output.

$$
\begin{cases}
\text{Residual} & r_t = x_t - (\beta_0 t + \beta_1) \\
\text{S.D.} & \sigma = \sqrt{\frac{1}{T}\sum_{t=1}^T r_t^2} \\
\text{Output} & \tilde{x}_t = r_t / \sigma
\end{cases}
$$

To achieve constant time and space complexity during inference, the linear trend needs to be replaced with a Recursive Moving Average (RMA) trend.

$$
\begin{cases}
\text{RMA Trend} & \mu_t = \alpha \mu_{t-1} + (1-\alpha)x_t \\

\text{Residual} & r_t = x_t - \mu_t \\

\text{S.D.} & \sigma_t = \sqrt{\alpha \sigma_{t-1}^2 + (1-\alpha)r_t^2} \\
\text{Output} & \tilde{x}_t = r_t / \sigma_t
\end{cases}
$$

By updating the trend using RMA, there is no need to cache historical states, ensuring constant time and memory complexity. Moreover, RMA exhibits a strong similarity to the linear trend, making it an effective alternative for real-time processing. The performance differences between the linear trend (chunk-based computation) and the RMA trend (frame-by-frame computation) are detailed in Table~\ref{tab:chunk_flow}.

\begin{table*}[ht]
\centering
\caption{5-Fold Intra-dataset results}
\label{tab:intra}
\begin{tabular}{lcccccccccccc}
\toprule
\multirow{2}{*}{Model} & \multicolumn{3}{c}{MMPD~\cite{tang2023mmpd}} & \multicolumn{3}{c}{VitalVideo~\cite{Toye2023VitalVA}} & \multicolumn{3}{c}{PURE~\cite{pure}} & \multicolumn{3}{c}{UBFC~\cite{ubfcrppg}} \\
\cmidrule(lr){2-4} \cmidrule(lr){5-7} \cmidrule(lr){8-10} \cmidrule(lr){11-13}
 & MAE & RMSE & R & MAE & RMSE & R & MAE & RMSE & R & MAE & RMSE & R \\
\midrule
EFFPhys~\cite{efficientphys} & 14.80 & 20.30 & 0.31 & 2.35 & 7.17 & 0.83 & 1.63 & 3.12 & 0.98 & 3.41 & 6.55 & 0.89 \\
TSCAN~\cite{mttscan} & 14.20 & 19.40 & 0.39 & 2.93 & 7.54 & 0.83 & 2.06 & 3.95 & 0.96 & 2.36 & 4.08 & 0.98 \\
PhysNet~\cite{physnet} & 8.13 & 12.40 & 0.56 & 0.69 & 2.66 & 0.98 & 0.93 & 2.08 & 0.99 & 1.76 & 2.92 & 0.98 \\
PhysFormer~\cite{physformer} & 9.47 & 14.00 & 0.47 & 0.66 & 2.76 & 0.98 & 2.66 & 5.32 & 0.90 & 5.42 & 8.99 & 0.85 \\
RhyMamba~\cite{zou2024rhythmmamba} & 8.21 & 12.00 & 0.61 & 0.83 & 3.69 & 0.96 & 0.26 & 0.53 & 1.00 & \textbf{0.53} & \textbf{0.73} & \textbf{1.00} \\
ME-rPPG & \textbf{5.58} & \textbf{9.84} & \textbf{0.76} & \textbf{0.64} & \textbf{2.66} & \textbf{0.98} & \textbf{0.26} & \textbf{0.53} & \textbf{1.00} & 0.68 & 1.16 & 1.00 \\
\bottomrule
\end{tabular}
\scriptsize \noindent \\MAE $\downarrow$ = Mean Absolute Error in HR estimation (Beats/Min). RMSE $\downarrow$ = Root Mean Squared Error in HR estimation (Beats/Min). R $\uparrow$ = Pearson Correlation in HR estimation.
\end{table*}

\subsection{State Space Duality}
SSMs are memory-efficient during inference. A fundamental SSM is defined as follows in Equation \eqref{eq:state space}, where $\mathbf{A} \in \mathbb{R}^{N×N}$ is the continuous state transition matrix, and $\mathbf{B} \in \mathbb{R}^{N \times p}$ and $\mathbf{C} \in \mathbb{R}^{q \times N}$ are the input and output projection matrices, respectively.

\begin{equation}
\begin{aligned}
h'(t) &= \mathbf{A}h(t) + \mathbf{B}x(t) \\
y(t) &= \mathbf{C}h(t)
\end{aligned}
\label{eq:state space}
\end{equation}

Using the Zero-Order Hold (ZOH) discretization method, any Linear Time-Invariant (LTI) system can be discretized as follows: $\bar{A} = e^{A \Delta t}$ and $\bar{B} = \left( \int_0^{\Delta t} e^{A \tau} d\tau \right) B$. This property ensures the equivalence of state transitions between the continuous and discrete systems, forming the core theorem of SSD.

According to Mamba-2~\cite{dao2024transformers}, the recursive relationship of the hidden state $h_t$ is transformed into matrix multiplication via global convolution expansion: $h_t = \sum_{k=1}^t \bar{A}^{t-k} \bar{B} x_k$. By constraining matrix $\bar{A}$ to be a diagonal matrix, matrix multiplication and inversion can be avoided, significantly improving training efficiency. The global hidden state $H$, composed of hidden states from all time steps, is expressed as \eqref{eq:global state}. Here, $L$ is a lower triangular matrix, whose elements are defined as $L_{i,j} = \bar{A}^{i-j}$ when $i \geq j$, and $\odot$ denotes the Hadamard product.

\begin{equation}
H = (L \odot (C \bar{B})) \cdot X
\label{eq:global state}
\end{equation} 

This formulation is equivalent to linear attention, expressed as $(L \odot (Q K^T)) \cdot V$. Thus, SSD exhibits long-range modeling capabilities similar to those of Transformers.
Due to the step-by-step inference characteristics of SSMs, linear complexity can be achieved during inference, enhancing computational efficiency and reducing latency.

\begin{table*}[ht]
\centering
\caption{Cross-dataset results}
\label{tab:cross}
\begin{tabular}{lcccccccccccccccc}
\toprule
\multirow{2}{*}{Model} & {Test Set} & \multicolumn{3}{c}{MMPD~\cite{tang2023mmpd}} & \multicolumn{3}{c}{VitalVideo~\cite{Toye2023VitalVA}} & \multicolumn{3}{c}{PURE~\cite{pure}} & \multicolumn{3}{c}{UBFC~\cite{ubfcrppg}} \\
\cmidrule(lr){2-2} \cmidrule(lr){3-5} \cmidrule(lr){6-8} \cmidrule(lr){9-11} \cmidrule(lr){12-14}
 & {Train Set}& MAE & RMSE & R & MAE & RMSE & R & MAE & RMSE & R & MAE & RMSE & R \\
\midrule
\multirow{2}{*}{EFFPhys} & RLAP & 9.18 & 13.70 & 0.56 & 6.75 & 19.30 & 0.51 & 2.48 & 5.48 & 0.98 & 0.81 & 1.71 & 1.00 \\
 & PURE & 17.60 & 23.00 & 0.26 & 3.83 & 8.42 & 0.79 & - & - & - & 2.96 & 5.11 & 0.97 \\
\multirow{2}{*}{TSCAN} & RLAP & 10.19 & 15.20 & 0.47 & 4.20 & 8.86 & 0.78 & 3.90 & 7.05 & 0.96 & 0.90 & 1.66 & 1.00 \\
 & PURE & 18.10 & 23.80 & 0.22 & 3.24 & 8.03 & 0.82 & - & - & - & 2.71 & 5.70 & 0.95 \\
\multirow{2}{*}{PhysNet} & RLAP & 11.20 & 16.40 & 0.42 & 0.89 & 3.56 & 0.96 & 0.63 & 2.34 & 1.00 & 0.66 & 1.03 & 1.00 \\
 & PURE & 15.10 & 20.50 & 0.29 & 1.32 & 5.13 & 0.93 & - & - & - & 1.62 & 3.49 & 0.98 \\
\multirow{2}{*}{PhysFormer} & RLAP & 12.80 & 18.90 & 0.28 & 4.97 & 10.10 & 0.70 & 0.91 & 3.32 & 0.99 & 0.50 & 0.75 & 1.00 \\
 & PURE & 13.00 & 18.10 & 0.44 & 1.08 & 4.07 & 0.95 & - & - & - & 1.15 & 2.06 & 0.99 \\
\multirow{2}{*}{RhyMamba} & RLAP & 9.62 & 14.90 & 0.47 & 2.85 & 8.91 & 0.80 & 1.63 & 4.55 & 0.98 & \textbf{0.44} & \textbf{0.68} & \textbf{1.00} \\
 & PURE & 10.90 & 15.20 & 0.42 & 1.74 & 5.87 & 0.91 & - & - & - & 0.63 & 1.07 & 1.00 \\
\multirow{2}{*}{ME-rPPG} & RLAP & \textbf{5.38} & \textbf{9.50} & \textbf{0.78} & \textbf{0.70} & \textbf{2.98} & \textbf{0.98} & \textbf{0.25} & \textbf{0.64} & \textbf{1.00} & 0.47 & 0.69 & 1.00 \\
 & PURE & 10.50 & 15.70 & 0.54 & 0.89 & 3.61 & 0.96 & - & - & - & 0.53 & 0.85 & 1.00 \\
\bottomrule
\end{tabular}
\scriptsize \noindent \\MAE $\downarrow$ = Mean Absolute Error in HR estimation (Beats/Min). RMSE $\downarrow$ = Root Mean Squared Error in HR estimation (Beats/Min). R $\uparrow$ = Pearson Correlation in HR estimation.
\end{table*}
\section{Experiments}
\label{sec:experiment}
In this section, we first introduce the datasets in Section~\ref{sec:dataset} and the implementation details in Section~\ref{sec:implementation}. We then compare our method with the SOTA methods in Section~\ref{sec:comparison}. We conduct an ablation study in Section~\ref{sec:ablation} to assess the impact of each component and the chunk length of training and test in our method. Finally, we validate our method in real-world scenarios in Section~\ref{sec:realworld}.

\subsection{Datasets}
\label{sec:dataset}
We use five benchmark datasets to evaluate our method: UBFC-rPPG~\cite{ubfcrppg}, PURE~\cite{pure}, MMPD~\cite{tang2023mmpd}, RLAP-rPPG~\cite{wang2023physbench}, and VitalVideo~\cite{Toye2023VitalVA}. UBFC-rPPG consists of 43 RGB videos recorded at 30 fps, capturing facial recordings under controlled indoor lighting. PURE includes 59 videos from 10 subjects with various head movements to test robustness against motion artifacts. MMPD contains 660 videos from 33 subjects under different lighting conditions and activities, providing diverse real-world scenarios. RLAP features recordings from 58 subjects with high synchronization between physiological and video data. VitalVideo, the largest dataset, includes recordings from 893 subjects with diverse skin tones and varied indoor lighting conditions, ensuring robustness against environmental variations. These datasets were selected for their accessibility, reliability, diversity, and synchronization quality, making them suitable for evaluating remote physiological signal extraction methods.

\subsection{Implementation Details}
\label{sec:implementation}
We have reproduced all baseline models according to the original text, where our model employs the Adam optimizer and Mean Squared Error (MSE) loss, trained for 20 epochs with a batch size of 32, and other parameters set to default. To mitigate the issue of insufficient high and low heart rate samples in the dataset, we generated an additional 30\% of data by time-scaling the training set, with a heart rate distribution ranging from 30 to 180 BPM. To ensure consistency in preprocessing and postprocessing, we defaulted to testing ME-rPPG in chunk mode.
All experiments were conducted with this platform: Ubuntu 22.04, CUDA 12.6, JAX 0.4.26. The hardware platform consisted of 2-way Xeon E5-2650v4 24-core, 4-way Tesla V100 Nvlink, 512G RAM. Training utilized FP16 mixed precision acceleration and Distributed Data Parallel (DDP). 


For user experiments, the model was exported using ONNX, and inference was performed on a consumer-grade laptop (Legion Y7000P, Intel i7-13620H, 16GB RAM), running our self-developed rPPG software, utilizing only the CPU for inference.

\subsection{Comparison with State-of-the-Art Methods}
\label{sec:comparison}
\textbf{Intra-dataset Experiments.} We first compare our method with five SOTA methods (Efficient-Phys~\cite{efficientphys}, TSCAN~\cite{mttscan}, PhysNet\cite{physnet}, PhysFormer~\cite{physformer}, RhythmMamba~\cite{zou2024rhythmmamba}) on four representative datasets (MMPD~\cite{tang2023mmpd}, VitalVideo~\cite{Toye2023VitalVA}, PURE~\cite{pure}, and UBFC~\cite{ubfcrppg}) using 5-fold subject-independent intra-dataset settings. The results are shown in Table~\ref{tab:intra}. Our method outperforms all other methods on almost all datasets, demonstrating its effectiveness in remote PPG and HR estimation. Specifically, ME-rPPG achieves the lowest MAE of 5.58 on MMPD and the highest Pearson Correlation of 1.00 on PURE and UBFC datasets, highlighting its superior accuracy and reliability.

\textbf{Cross-dataset Experiments.} To assess the generalization ability of our method, we evaluate its performance in cross-dataset settings. The results in Table~\ref{tab:cross} indicate that ME-rPPG achieves the best performance across most cross-dataset settings, demonstrating strong generalization. Specifically, ME-rPPG achieves the lowest MAE of 5.38 and the highest R of 0.78 on the challenging MMPD dataset when trained on the RLAP dataset. Additionally, ME-rPPG outperforms all methods on the largest VitalVideo dataset (MAE: 0.70, R: 0.98) and the PURE dataset (MAE: 0.25, R: 1.00), confirming its robustness and reliability. Compared to intra-dataset results, ME-rPPG maintains consistent performance across different datasets, whereas other methods show significant performance drops. This demonstrates our method's effectiveness in handling diverse real-world scenarios and generalizing well across datasets.

\textbf{Computational Efficiency.} We further assess the computational efficiency of our method by comparing its inference time against other SOTA approaches. As shown in Table~\ref{tab:efficiency}, ME-rPPG achieves the best trade-off between accuracy and efficiency, underscoring its practicality for real-world applications. ME-Flow achieves the lowest latency of 9.46 ms, which is even lower than the frame capture time of 33.3 ms at 30 fps, making it suitable for real-time PPG estimation on resource-constrained devices.

\begin{table}[tp]
\centering
\caption{Ablation Study on Training and Test Chunk Length}
\label{tab:chunk}
\begin{tabular}{cccccc}
\toprule
Train\textbackslash Test & 160 & 320 & 640 & 1280 & 1800 \\
\midrule
160 & 6.13 & 6.05 & 6.11 & 6.15 & 6.11 \\
320 & 6.17 & 6.19 & 6.17 & 6.09 & 5.78 \\
640 & 6.24 & 6.05 & 6.02 & 5.86 & 5.79 \\
1280 & 5.92 & 5.91 & 5.80 & 5.66 & 5.75 \\
1600 & 5.93 & 5.73 & 5.51 & 5.42 & \textbf{5.38} \\
\bottomrule
\end{tabular}
\scriptsize \noindent \\ Experiments are trained on RLAP and testes on MMPD. MAE is calculated in HR estimation (Beats/Min). Each video is divided into chunks of different lengths.
\end {table}

\subsection{Ablation Study}
\label{sec:ablation}


\textbf{Ablation Study on Components.} We conduct an ablation study to validate the effectiveness of each component in ME-rPPG. We compare the performance after removing different components, including TN and SSD modules. The results in Table~\ref{tab:ablation} show that our method performs best when all components are included. The TN module maintains the stability of temporal features, while the SSD module provides attention to assign weights to different frames. Combining both enhances our method's ability to handle challenging real-world scenarios, demonstrating the necessity of each module.

\begin{table*}[ht]
\centering
\caption{Ablation Study on Components}
\label{tab:ablation}
\begin{tabular}{lcccccccccccc}
\toprule
\multirow{2}{*}{Model} & \multicolumn{3}{c}{MMPD} & \multicolumn{3}{c}{VitalVideo} & \multicolumn{3}{c}{PURE} & \multicolumn{3}{c}{UBFC} \\
\cmidrule(lr){2-4} \cmidrule(lr){5-7} \cmidrule(lr){8-10} \cmidrule(lr){11-13}
 & MAE & RMSE & R & MAE & RMSE & R & MAE & RMSE & R & MAE & RMSE & R \\
\midrule
w/o TN & 11.60 & 17.60 & 0.43 & 0.88 & 3.45 & 0.97 & 0.80 & 3.97 & 0.99 & 0.52 & 0.73 & 1.00 \\
w/o SSD & 10.80 & 16.70 & 0.51 & 1.10 & 3.80 & 0.96 & 0.34 & 0.87 & 1.00 & 0.63 & 0.85 & 1.00 \\
ME-rPPG & \textbf{5.38} & \textbf{9.50} & \textbf{0.78} & \textbf{0.70} & \textbf{2.98} & \textbf{0.98} & \textbf{0.25} & \textbf{0.64} & \textbf{1.00} & \textbf{0.47} & \textbf{0.69} & \textbf{1.00} \\
\bottomrule
\end{tabular}
\scriptsize \noindent \\MAE $\downarrow$ = Mean Absolute Error in HR estimation (Beats/Min). RMSE $\downarrow$ = Root Mean Squared Error in HR estimation (Beats/Min). R $\uparrow$ = Pearson Correlation in HR estimation.
\end{table*}

\textbf{Ablation Study on Training and Test Chunk Length.} We also examine the impact of different training and test chunk lengths on model performance. The results in Table~\ref{tab:chunk} show that longer chunks lead to better performance by capturing more temporal information. This finding aligns with previous studies~\cite{zou2024rhythmmamba} that longer test lengths can improve rPPG performance. With our memory-efficient design, we can train on entire videos, further proving that longer training lengths also enhance performance. Our method achieves the best results with a training chunk length of 1600 and a test chunk length of 1800, demonstrating its effectiveness in handling long-term temporal dependencies.

\subsection{Validation in Real-world Scenarios}
\label{sec:realworld}

To validate our method in real-world scenarios, we conducted a real-time experiment using a self-developed application equipped with ME-Flow and PhysNet. Approved by the Institutional Review Board (IRB), we recruited 11 subjects (7 males, 4 females, Age: 22.46 $\pm$ 4.03). Each subject sat 40-50 cm from a laptop under various indoor lighting conditions (day and night) and performed both static and moving head scenarios. The system recorded facial videos using a Logitech C920 at 30 fps and displayed the predicted PPG waveforms in real time. Ground-truth physiological signals were recorded using a Contec CMS50E oximeter. We compared the heart rate estimated by our method with the ground truth. The results in Table~\ref{tab:realworld} show that ME-Flow achieves high accuracy, with a 19.5\% improvement in static scenarios and a 49.7\% improvement in moving scenarios (compared with PhysNet~\cite{physnet}).

\begin{table*}[ht]
\centering
\caption{Real-time User Evaluation}
\label{tab:realworld}
\begin{tabular}{lcccccccccc}
\toprule
\multirow{2}{*}{Method} & \multicolumn{3}{c}{Static} & \multicolumn{3}{c}{Moving} & \multicolumn{4}{c}{User Evaluation} \\
\cmidrule(lr){2-4} \cmidrule(lr){5-7} \cmidrule(lr){8-11}
 & MAE & RMSE & R & MAE & RMSE & R & Overall & Fluency & Accuracy & Stability \\
\midrule
PhysNet & 2.21 & 2.47 & 0.98 & 5.03 & 12.30 & 0.53 & 4.00 & 3.64 & 4.73 & 4.55 \\
ME-rPPG & \textbf{1.78} & \textbf{2.29} & \textbf{0.99} & \textbf{2.53} & \textbf{3.06} & \textbf{0.98} & \textbf{5.73} & \textbf{6.09} & \textbf{5.55} & \textbf{5.36} \\
\bottomrule
\end{tabular}
\scriptsize \noindent \\MAE $\downarrow$ = Mean Absolute Error in HR estimation (Beats/Min). RMSE $\downarrow$ = Root Mean Squared Error in HR estimation (Beats/Min). R $\uparrow$ = Pearson Correlation in HR estimation.
\end{table*}

\begin{figure}[t]
\centering
\includegraphics[width=0.99\linewidth]{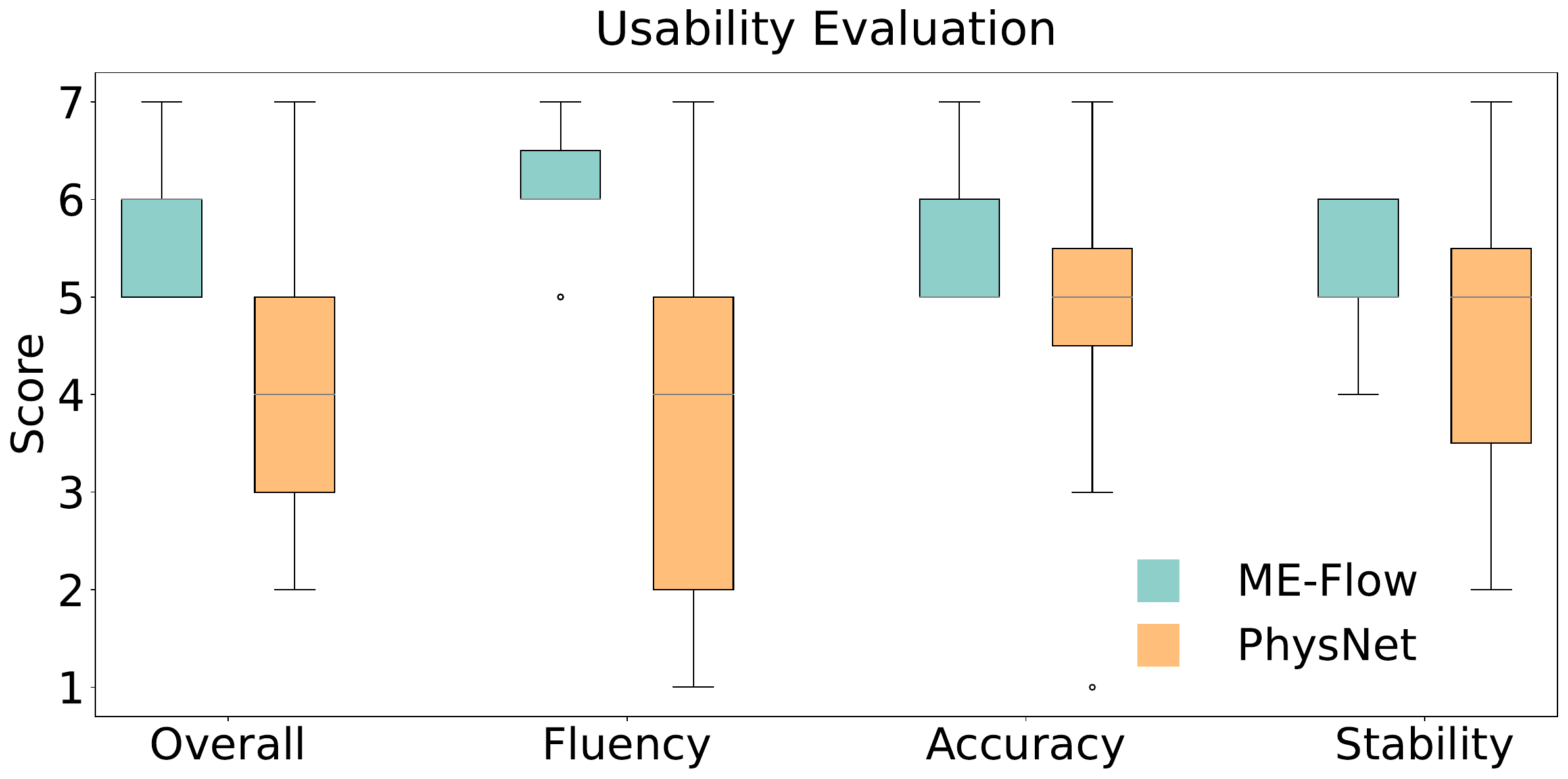}
\caption{\textbf{User Experience Evaluation.} Users scored the overall experience, fluency, accuracy and stability subjectively.}
\label{fig:evaluation}
\end{figure}

User feedback was collected using a 7-point Likert~\cite{joshi2015likert} scale (higher is better) to evaluate the user experience of our application. Participants do not frequently use health-related devices (Score: 2.82 ± 2.14), yet they express a high willingness to adopt the application (Score: 5.18 ± 1.33). They are satisfied with our rPPG application (Score: 5.09 $\pm$ 1.14) and are willing to use it in the future. The results in Figure~\ref{fig:evaluation} indicate that users rate our system higher in overall performance, fluency, accuracy, and stability compared to the PhysNet method (p $<$ 0.05, Wilcoxon test), demonstrating its effectiveness in real-world scenarios.

\begin{table}[ht]
\centering
\caption{Performance comparison of ME-Chunk and ME-Flow.}
\label{tab:chunk_flow}
\begin{tabular}{ccccc}
\toprule
Inference & MMPD & VitalVideo & PURE & UBFC \\ \midrule
ME-Chunk & 5.38 & 0.70 & 0.25 & 0.47 \\
ME-Flow & 6.94 & 0.85 & 0.38 & 0.52 \\ 
\bottomrule
\end{tabular}
\scriptsize \noindent \\MAE $\downarrow$ = Mean Absolute Error in HR estimation (Beats / Min).
\end{table}

\section{Discussion}
\label{sec:discussion}

\subsection{Effective and Generalizable ME-rPPG}

ME-rPPG exhibits stable performance in intra-dataset evaluations. While ME-rPPG underperforms RhyMamba on UBFC, it excels on MMPD and the large-scale VitalVideo dataset. In cross-dataset evaluations, ME-rPPG shows strong generalization, maintaining consistent performance, whereas other methods experience significant degradation. This indicates that ME-rPPG is more suitable for real-world applications. During cross-dataset validation, whether trained on RLAP or PURE, the MAE on VitalVideo and UBFC remains below 1 BPM, achieving medical-grade accuracy. Additionally, on the MMPD dataset, which closely resembles real-world mobile scenarios, the lowest error achieved is 5.38 BPM, demonstrating its potential in practical applications.

The robust generalization of ME-rPPG can be attributed to its TN and TSD architectural design. This design captures rPPG features across both temporal and spatial dimensions while reducing noise interference. It supports longer inputs and outputs, allowing training on extended video segments that contain long-term periodic features consistent shared across various datasets. Previous methods focused on short video segments, emphasizing short-term features that vary with recording environments and devices. Studies~\cite{physnet} found that longer videos did not benefit CNN architectures due to limited receptive fields. As analyzed in Table~\ref{tab:memory complexity}, increasing the input length for Transformer architectures results in a quadratic increase in computational cost, which is impractical for large-scale training and inference. In contrast, ME-rPPG efficiently captures long-term dependencies with minimal computational overhead, allowing training on extended video segments and significantly enhancing cross-dataset generalization.

\subsection{Efficient and Real-time ME-rPPG}

Real-time performance is crucial for monitoring in medical and video streaming applications. Our user study indicates that the fluency and stability of rPPG systems significantly impact user experience and trust. Traditional methods like Green~\cite{verkruysse2008green} use the average of the green channel of a single frame for rPPG prediction but suffer from motion artifacts. Subsequent deep learning methods improve performance but require long inputs, limiting real-time applications. While models like RTrPPG~\cite{botina2022rtrppg} claim millisecond-level inference times, they still require pre-recorded input, preventing true streaming-based rPPG output. TS-CAN , which achieves performance closest to our application, requires users to record for 30 seconds before obtaining pulse waveform predictions\footnote{\url{https://vitals.cs.washington.edu/}}. With only 580K parameters, ME-rPPG uses 3.6 MB of memory and has a latency of 9.46 ms, making real-time rPPG prediction feasible on mobile devices. To the best of our knowledge, ME-rPPG is the first model to enable real-time rPPG waveform prediction while recording.

The real-time efficiency of ME-rPPG is mainly due to the transferable state space in our TSD architecture. ME-rPPG uses a transferable state space, storing forward information rather than constructing frame-to-frame dependencies within fixed video segments. This allows each frame to generate real-time BVP predictions and accumulate forward information for more stable predictions. As shown in Table~\ref{tab:chunk_flow}, ME-rPPG’s frame-wise inference exhibits slightly higher error than chunk-wise inference, primarily due to trend calculation limitations in the TN module. Frame-wise inference relies on sliding averages rather than linear regression, as discussed in Section~\ref{sec:method}. This difference may cause ME-rPPG to struggle with noise in the MMPD dataset but still maintain good performance on static datasets such as VitalVideo, as shown in Figure~\ref{fig:waveform}. The lightweight TN module and TSD design also contribute to ME-rPPG's real-time performance, making it suitable for time-sensitive and resource-limited applications.

\subsection{Limitations and Future Work}

Although ME-rPPG has made significant progress in performance and real-time capabilities, certain limitations remain. First, to ensure unidirectional state space transfer, our model cannot access backward information during training, potentially missing long-term periodic features. Second, although we validated our model on datasets with more than 1000 subjects, we have not yet tested it in real clinical scenarios, where performance for users with cardiovascular diseases remains to be evaluated. Finally, to maximize training and testing efficiency, we used the efficient Jax platform for all experiments instead of the more widely used PyTorch platform, which may lead to some differences.

Future work can extend ME-rPPG to predict additional cardiovascular signals from facial videos, including blood oxygen~\cite{tang2025camerameasurementbloodoxygen} and blood pressure~\cite{curran2023camera}. Enhancing rPPG accuracy through multimodal fusion with thermal imaging or IMU sensors is another potential direction. Optimizing computational efficiency will enable real-time processing at higher frame rates on lightweight hardware, expanding its practical applications in remote healthcare monitoring.

\section{Conclusion}
\label{sec:conclusion}

In this paper, we propose ME-rPPG, a memory-efficient rPPG algorithm based on temporal-spatial state space duality, designed for arbitrary length video training and inference. Our approach effectively captures subtle temporal changes and spatial representations of facial vascular information, achieving significant improvements over state-of-the-art methods on large-scale datasets such as MMPD and VitalVideo. ME-rPPG demonstrates superior accuracy and memory efficiency, with improvements of up to 60.3\% in accuracy and 98.4\% in memory efficiency. To the best of our knowledge, ME-rPPG is the first rPPG model capable of real-time inference from single-frame input, reducing computational load and achieving better user experience in real-world applications. Our method provides new directions for future research on rPPG, effectively balancing accuracy and efficiency for practical deployment.

{
    \small
    \bibliographystyle{ieeenat_fullname}
    \bibliography{main}

\begin{thebibliography}{45}
\providecommand{\natexlab}[1]{#1}
\providecommand{\url}[1]{\texttt{#1}}
\expandafter\ifx\csname urlstyle\endcsname\relax
  \providecommand{\doi}[1]{doi: #1}\else
  \providecommand{\doi}{doi: \begingroup \urlstyle{rm}\Url}\fi

\bibitem[Bobbia et~al.(2019)Bobbia, Macwan, Benezeth, Mansouri, and
  Dubois]{ubfcrppg}
Serge Bobbia, Richard Macwan, Yannick Benezeth, Alamin Mansouri, and Julien
  Dubois.
\newblock Unsupervised skin tissue segmentation for remote
  photoplethysmography.
\newblock \emph{Pattern Recognition Letters}, 124:\penalty0 82--90, 2019.

\bibitem[Botina-Monsalve et~al.(2022{\natexlab{a}})Botina-Monsalve, Benezeth,
  and Miteran]{botina2022rtrppg}
Deivid Botina-Monsalve, Yannick Benezeth, and Johel Miteran.
\newblock Rtrppg: An ultra light 3dcnn for real-time remote
  photoplethysmography.
\newblock In \emph{Proceedings of the IEEE/CVF Conference on Computer Vision
  and Pattern Recognition}, pages 2146--2154, 2022{\natexlab{a}}.

\bibitem[Botina-Monsalve et~al.(2022{\natexlab{b}})Botina-Monsalve, Benezeth,
  and Miteran]{rtrppg}
Deivid Botina-Monsalve, Yannick Benezeth, and Johel Miteran.
\newblock Rtrppg: An ultra light 3dcnn for real-time remote
  photoplethysmography.
\newblock In \emph{Proceedings of the IEEE/CVF Conference on Computer Vision
  and Pattern Recognition (CVPR) Workshops}, pages 2146--2154,
  2022{\natexlab{b}}.

\bibitem[Chen and McDuff(2018)]{chen2018deepphys}
Weixuan Chen and Daniel McDuff.
\newblock Deepphys: Video-based physiological measurement using convolutional
  attention networks.
\newblock In \emph{Proceedings of the European Conference on Computer Vision
  (ECCV)}, pages 349--365, 2018.

\bibitem[Curran et~al.(2023)Curran, Liu, McDuff, Patel, and
  Yang]{curran2023camera}
Theodore Curran, Xin Liu, Daniel McDuff, Shwetak Patel, and Eugene Yang.
\newblock Camera-based remote photoplethysmography to measure heart rate and
  blood pressure in ambulatory patients with cardiovascular disease:
  Preliminary analysis.
\newblock \emph{Journal of the American College of Cardiology}, 81\penalty0
  (8\_Supplement):\penalty0 2301--2301, 2023.

\bibitem[Dao and Gu(2024)]{dao2024transformers}
Tri Dao and Albert Gu.
\newblock Transformers are {{SSMs}}: {{Generalized Models}} and {{Efficient
  Algorithms Through Structured State Space Duality}}.
\newblock \emph{ArXiv}, 2024.

\bibitem[Du et~al.(2022)Du, Zhang, Su, Wang, Teng, and Liu]{du2022multimodal}
Guanglong Du, Linlin Zhang, Kang Su, Xueqian Wang, Shaohua Teng, and Peter~X.
  Liu.
\newblock A {{Multimodal Fusion Fatigue Driving Detection Method Based}} on
  {{Heart Rate}} and {{PERCLOS}}.
\newblock \emph{IEEE Transactions on Intelligent Transportation Systems},
  23\penalty0 (11):\penalty0 21810--21820, 2022.

\bibitem[Gideon and Stent(2021)]{Gideon2021TheWT}
John Gideon and Simon Stent.
\newblock The way to my heart is through contrastive learning: Remote
  photoplethysmography from unlabelled video.
\newblock \emph{2021 IEEE/CVF International Conference on Computer Vision
  (ICCV)}, pages 3975--3984, 2021.

\bibitem[Gu and Dao(2023)]{gu2023mamba}
Albert Gu and Tri Dao.
\newblock Mamba: {{Linear-Time Sequence Modeling}} with {{Selective State
  Spaces}}.
\newblock \emph{ArXiv}, 2023.

\bibitem[He et~al.(2019)He, Pham, Valliappan, Xu, Roberts, Lagun, and
  Navalpakkam]{he2019device}
Junfeng He, Khoi Pham, Nachiappan Valliappan, Pingmei Xu, Chase Roberts, Dmitry
  Lagun, and Vidhya Navalpakkam.
\newblock On-device few-shot personalization for real-time gaze estimation.
\newblock In \emph{Proceedings of the IEEE International Conference on Computer
  Vision Workshops}, pages 0--0, 2019.

\bibitem[Joshi et~al.(2015)Joshi, Kale, Chandel, and Pal]{joshi2015likert}
Ankur Joshi, Saket Kale, Satish Chandel, and D~Kumar Pal.
\newblock Likert scale: Explored and explained.
\newblock \emph{British journal of applied science \& technology}, 7\penalty0
  (4):\penalty0 396, 2015.

\bibitem[Joshi and Cho(2024)]{joshi2024ibvp}
Jitesh Joshi and Youngjun Cho.
\newblock Ibvp dataset: Rgb-thermal rppg dataset with high resolution signal
  quality labels.
\newblock \emph{Electronics}, 13\penalty0 (7):\penalty0 1334, 2024.

\bibitem[Kuang et~al.(2023)Kuang, Ao, Ma, and Liu]{kuang2023shuffle}
Hailan Kuang, Can Ao, Xiaolin Ma, and Xinhua Liu.
\newblock Shuffle-rppgnet: Efficient network with global context for remote
  heart rate variability measurement.
\newblock \emph{IEEE Sensors Journal}, 2023.

\bibitem[Li et~al.(2024)Li, Li, Wang, He, Wang, Wang, and
  Qiao]{li2024videomamba}
Kunchang Li, Xinhao Li, Yi Wang, Yinan He, Yali Wang, Limin Wang, and Yu Qiao.
\newblock {{VideoMamba}}: {{State Space Model}} for {{Efficient Video
  Understanding}}.
\newblock arXiv, 2024.

\bibitem[Li and Yin(2023)]{Li2023ContactlessPE}
Zhihua Li and Lijun Yin.
\newblock Contactless pulse estimation leveraging pseudo labels and
  self-supervision.
\newblock \emph{2023 IEEE/CVF International Conference on Computer Vision
  (ICCV)}, pages 20531--20540, 2023.

\bibitem[Lin et~al.(2019)Lin, Gan, and Han]{lin2019tsm}
Ji Lin, Chuang Gan, and Song Han.
\newblock Tsm: Temporal shift module for efficient video understanding.
\newblock In \emph{Proceedings of the IEEE/CVF International Conference on
  Computer Vision}, pages 7083--7093, 2019.

\bibitem[Liu et~al.(2024{\natexlab{a}})Liu, Tang, Jiang, Wang, Liu, Li, and
  Shi]{liu2024summit}
Ke Liu, Jiankai Tang, Zhang Jiang, Yuntao Wang, Xiaojing Liu, Dong Li, and
  Yuanchun Shi.
\newblock Summit vitals: Multi-camera and multi-signal biosensing at high
  altitudes.
\newblock In \emph{The 21st IEEE International Conference on Ubiquitous
  Intelligence and Computing (UIC 2024)}, 2024{\natexlab{a}}.

\bibitem[Liu et~al.(2024{\natexlab{b}})Liu, Tang, Li, Qi, Li, Wang, Wang, and
  Chen]{Liu2024SpikingPhysFormerCR}
Mingxuan Liu, Jiankai Tang, Haoxiang Li, Jiahao Qi, Siwei Li, Kegang Wang,
  Yuntao Wang, and Hong Chen.
\newblock Spiking-physformer: Camera-based remote photoplethysmography with
  parallel spike-driven transformer.
\newblock \emph{ArXiv}, abs/2402.04798, 2024{\natexlab{b}}.

\bibitem[Liu et~al.(2020{\natexlab{a}})Liu, Fromm, Patel, and
  McDuff]{liu2020multi}
Xin Liu, Josh Fromm, Shwetak Patel, and Daniel McDuff.
\newblock Multi-task temporal shift attention networks for on-device
  contactless vitals measurement.
\newblock \emph{Advances in Neural Information Processing Systems},
  33:\penalty0 19400--19411, 2020{\natexlab{a}}.

\bibitem[Liu et~al.(2020{\natexlab{b}})Liu, Fromm, Patel, and McDuff]{mttscan}
Xin Liu, Josh Fromm, Shwetak Patel, and Daniel McDuff.
\newblock Multi-task temporal shift attention networks for on-device
  contactless vitals measurement.
\newblock In \emph{Advances in Neural Information Processing Systems}, pages
  19400--19411. Curran Associates, Inc., 2020{\natexlab{b}}.

\bibitem[Liu et~al.(2021)Liu, Hill, Jiang, Patel, and McDuff]{efficientphys}
Xin Liu, Brian~L. Hill, Ziheng Jiang, Shwetak Patel, and Daniel McDuff.
\newblock Efficientphys: Enabling simple, fast and accurate camera-based vitals
  measurement, 2021.

\bibitem[Liu et~al.(2023)Liu, Narayanswamy, Paruchuri, Zhang, Tang, Zhang,
  Sengupta, Patel, Wang, and McDuff]{liu2022rppgtoolbox}
Xin Liu, Girish Narayanswamy, Akshay Paruchuri, Xiaoyu Zhang, Jiankai Tang,
  Yuzhe Zhang, Roni Sengupta, Shwetak Patel, Yuntao Wang, and Daniel McDuff.
\newblock rppg-toolbox: Deep remote ppg toolbox.
\newblock In \emph{Advances in Neural Information Processing Systems}, pages
  68485--68510. Curran Associates, Inc., 2023.

\bibitem[Lu et~al.(2021)Lu, Han, and Zhou]{lu2021dual}
Hao Lu, Hu Han, and S~Kevin Zhou.
\newblock Dual-gan: Joint bvp and noise modeling for remote physiological
  measurement.
\newblock In \emph{Proceedings of the IEEE/CVF Conference on Computer Vision
  and Pattern Recognition}, pages 12404--12413, 2021.

\bibitem[Luo et~al.(2024)Luo, Xie, and Yu]{luo2024physmamba}
Chaoqi Luo, Yiping Xie, and Zitong Yu.
\newblock Physmamba: Efficient remote physiological measurement with slowfast
  temporal difference mamba.
\newblock In \emph{Chinese Conference on Biometric Recognition}, pages
  248--259. Springer, 2024.

\bibitem[McDuff(2023)]{mcduff2023camera}
Daniel McDuff.
\newblock Camera measurement of physiological vital signs.
\newblock \emph{ACM Computing Surveys}, 55\penalty0 (9):\penalty0 1--40, 2023.

\bibitem[McDuff et~al.(2015)McDuff, Estepp, Piasecki, and
  Blackford]{mcduff2015survey}
Daniel~J McDuff, Justin~R Estepp, Alyssa~M Piasecki, and Ethan~B Blackford.
\newblock A survey of remote optical photoplethysmographic imaging methods.
\newblock In \emph{2015 37th annual international conference of the IEEE
  engineering in medicine and biology society (EMBC)}, pages 6398--6404. IEEE,
  2015.

\bibitem[Nowara et~al.(2020)Nowara, McDuff, and
  Veeraraghavan]{nowara_benefit_2020}
Ewa Nowara, Daniel McDuff, and Ashok Veeraraghavan.
\newblock The {Benefit} of {Distraction}: {Denoising} {Remote} {Vitals}
  {Measurements} using {Inverse} {Attention}.
\newblock \emph{arXiv:2010.07770 [cs, eess]}, 2020.
\newblock arXiv: 2010.07770.

\bibitem[Poh et~al.(2011)Poh, McDuff, and Picard]{poh2011medical}
Ming-Zher Poh, Daniel McDuff, and Rosalind Picard.
\newblock A medical mirror for non-contact health monitoring.
\newblock In \emph{ACM SIGGRAPH 2011 Emerging Technologies}, pages 1--1. 2011.

\bibitem[Song et~al.(2020)Song, Chen, Cheng, Li, Liu, and
  Chen]{song_pulsegan_2020}
Rencheng Song, Huan Chen, Juan Cheng, Chang Li, Yu Liu, and Xun Chen.
\newblock {PulseGAN}: {Learning} to generate realistic pulse waveforms in
  remote photoplethysmography.
\newblock \emph{arXiv:2006.02699 [eess]}, 2020.
\newblock arXiv: 2006.02699.

\bibitem[Stricker et~al.(2014)Stricker, Müller, and Gross]{pure}
Ronny Stricker, Steffen Müller, and Horst-Michael Gross.
\newblock Non-contact video-based pulse rate measurement on a mobile service
  robot.
\newblock In \emph{The 23rd IEEE International Symposium on Robot and Human
  Interactive Communication}, pages 1056--1062, 2014.

\bibitem[Tang et~al.(2023)Tang, Chen, Wang, Shi, Patel, McDuff, and
  Liu]{tang2023mmpd}
Jiankai Tang, Kequan Chen, Yuntao Wang, Yuanchun Shi, Shwetak Patel, Daniel
  McDuff, and Xin Liu.
\newblock Mmpd: Multi-domain mobile video physiology dataset, 2023.

\bibitem[Tang et~al.(2025)Tang, Liu, McDuff, Jiang, Hu, Zhou, Nagao, Suzuki,
  Nagahama, Li, Ji, Shi, Nishidate, and
  Wang]{tang2025camerameasurementbloodoxygen}
Jiankai Tang, Xin Liu, Daniel McDuff, Zhang Jiang, Hongming Hu, Luxi Zhou,
  Nodoka Nagao, Haruta Suzuki, Yuki Nagahama, Wei Li, Linhong Ji, Yuanchun Shi,
  Izumi Nishidate, and Yuntao Wang.
\newblock Camera measurement of blood oxygen saturation.
\newblock 2025.

\bibitem[Toye(2023)]{Toye2023VitalVA}
Pieter-Jan Toye.
\newblock Vital videos: A dataset of face videos with ppg and blood pressure
  ground truths.
\newblock 2023.

\bibitem[Verkruysse et~al.(2008)Verkruysse, Svaasand, and
  Nelson]{verkruysse2008green}
Wim Verkruysse, Lars~O Svaasand, and J~Stuart Nelson.
\newblock Remote plethysmographic imaging using ambient light.
\newblock \emph{Optics express}, 16\penalty0 (26):\penalty0 21434--21445, 2008.

\bibitem[Wang et~al.(2024{\natexlab{a}})Wang, Tang, Wei, Liu, Liu, and
  Wang]{wang2024plugandplaytemporalnormalizationmodule}
Kegang Wang, Jiankai Tang, Yantao Wei, Mingxuan Liu, Xin Liu, and Yuntao Wang.
\newblock A plug-and-play temporal normalization module for robust remote
  photoplethysmography, 2024{\natexlab{a}}.

\bibitem[Wang et~al.(2024{\natexlab{b}})Wang, Wei, Tang, Wang, Tong, Gao, Ma,
  and Zhao]{wang2023physbench}
Kegang Wang, Yantao Wei, Jiankai Tang, Yuntao Wang, Mingwen Tong, Jie Gao,
  Yujian Ma, and Zhongjin Zhao.
\newblock Camera-based hrv prediction for remote learning environments,
  2024{\natexlab{b}}.

\bibitem[Wang et~al.(2017)Wang, den Brinker, Stuijk, and de~Haan]{pos}
Wenjin Wang, Albertus~C. den Brinker, Sander Stuijk, and Gerard de Haan.
\newblock Algorithmic principles of remote ppg.
\newblock \emph{IEEE Transactions on Biomedical Engineering}, 64\penalty0
  (7):\penalty0 1479--1491, 2017.

\bibitem[Wu et~al.(2012)Wu, Rubinstein, Shih, Guttag, Durand, and
  Freeman]{wu2012eulerian}
Hao-Yu Wu, Michael Rubinstein, Eugene Shih, John Guttag, Fr{\'e}do Durand, and
  William Freeman.
\newblock Eulerian video magnification for revealing subtle changes in the
  world.
\newblock \emph{ACM transactions on graphics (TOG)}, 31\penalty0 (4):\penalty0
  1--8, 2012.

\bibitem[Wu et~al.(2025)Wu, Xie, Zhao, He, Luo, Deng, and
  Yu]{wu2025cardiacmamba}
Zheng Wu, Yiping Xie, Bo Zhao, Jiguang He, Fei Luo, Ning Deng, and Zitong Yu.
\newblock Cardiacmamba: A multimodal rgb-rf fusion framework with state space
  models for remote physiological measurement.
\newblock \emph{arXiv preprint arXiv:2502.13624}, 2025.

\bibitem[Xu et~al.(2024)Xu, Yang, Wang, Cai, Du, and Chen]{xu2024visual}
Rui Xu, Shu Yang, Yihui Wang, Yu Cai, Bo Du, and Hao Chen.
\newblock Visual {{Mamba}}: {{A Survey}} and {{New Outlooks}}, 2024.

\bibitem[Yu et~al.(2019)Yu, Li, and Zhao]{physnet}
Zitong Yu, Xiaobai Li, and Guoying Zhao.
\newblock Remote photoplethysmograph signal measurement from facial videos
  using spatio-temporal networks.
\newblock In \emph{Proc. BMVC}, 2019.

\bibitem[Yu et~al.(2022{\natexlab{a}})Yu, Shen, Shi, Zhao, Torr, and
  Zhao]{physformer}
Zitong Yu, Yuming Shen, Jingang Shi, Hengshuang Zhao, Philip Torr, and Guoying
  Zhao.
\newblock Physformer: Facial video-based physiological measurement with
  temporal difference transformer.
\newblock In \emph{CVPR}, 2022{\natexlab{a}}.

\bibitem[Yu et~al.(2022{\natexlab{b}})Yu, Shen, Shi, Zhao, Torr, and
  Zhao]{yu2022physformer}
Zitong Yu, Yuming Shen, Jingang Shi, Hengshuang Zhao, Philip~HS Torr, and
  Guoying Zhao.
\newblock Physformer: Facial video-based physiological measurement with
  temporal difference transformer.
\newblock In \emph{Proceedings of the IEEE/CVF conference on computer vision
  and pattern recognition}, pages 4186--4196, 2022{\natexlab{b}}.

\bibitem[Zou et~al.(2024{\natexlab{a}})Zou, Guo, Chen, and
  Ma]{Zou2024RhythmFormerER}
Bochao Zou, Zizheng Guo, Jiansheng Chen, and Huimin Ma.
\newblock Rhythmformer: Extracting rppg signals based on hierarchical temporal
  periodic transformer.
\newblock \emph{ArXiv}, abs/2402.12788, 2024{\natexlab{a}}.

\bibitem[Zou et~al.(2024{\natexlab{b}})Zou, Guo, Hu, and
  Ma]{zou2024rhythmmamba}
Bochao Zou, Zizheng Guo, Xiaocheng Hu, and Huimin Ma.
\newblock Rhythmmamba: Fast remote physiological measurement with arbitrary
  length videos.
\newblock \emph{arXiv preprint arXiv:2404.06483}, 2024{\natexlab{b}}.

\end{thebibliography}
}

\end{document}